\documentclass[letterpaper]{article} 
\usepackage{aaai2027} 
\usepackage[hyphens]{url}  
\usepackage{graphicx} 
\usepackage{natbib}  
\usepackage{caption} 
\usepackage{algorithm}
\usepackage{enumitem}
\usepackage[most]{tcolorbox}
\usepackage{algorithm}
\usepackage[noend]{algpseudocode}
\usepackage{todonotes}
\usepackage{amsmath}
\usepackage{amssymb}
\usepackage{newfloat}
\usepackage{listings}
\definecolor{hayden_bg}{RGB}{198, 224, 244}   
\definecolor{annie_bg}{RGB}{255, 218, 185}     
\DeclareCaptionStyle{ruled}{labelfont=normalfont,labelsep=colon,strut=off} 
\floatstyle{ruled}
\newfloat{listing}{tb}{lst}{}
\floatname{listing}{Listing}

\newcommand{\titleac}{REFLECT}

\usepackage{booktabs}

\title{REFLECT: Role-Aware Detection of\\ Sentiment, Emotion, Politeness, and Formality}

\author{
    Sangpil Youm,
    Annabel Stocks Natalias,
    Hayden Heninger,
    Leah Jones,
    Xiaoyang Jin,
    Xiaoya Zhang,
    Bonnie Dorr
}
\affiliations{
    University of Florida\\
    \{youms, astocksnatalias, haydenheninger, leahjones, xiaoyangjin, xiaoyazhang, bonniejdorr\}@ufl.edu
}

\begin{document}

\maketitle

\begin{abstract}

Detecting communicative meaning in dialogue requires modeling not only lexical content but also the participant-role structures through which sentiment, emotion, politeness, and formality (SEPF) are expressed. Existing approaches rely on contextual language representations while leaving semantic roles implicit. We introduce a role-aware transformer architecture that maps the four SEPF dimensions to theory-grounded Semantic Role Labeling (SRL) configurations, incorporating these configurations directly into the transformer's self-attention mechanism. Role-aware attention models non-directional internal states (sentiment and emotion), while directional role-aware attention captures initiator-to-recipient relations for external cues (politeness and formality). The resulting role-aware representations are fused with contextual embeddings through a learned gate that adaptively weights how much role information each input contributes, rather than appending semantic roles as an auxiliary feature vector. Across four benchmark datasets, the framework yields statistically reliable gains over strong BERT-based baselines for emotion, politeness, and formality, with gains concentrated in theoretically predicted cases such as low-frequency, complex emotions and highly polite expressions; sentiment improvements are marginal and non-significant, suggesting that sentiment is conveyed more through lexical choice than participant-role structure. Additional evaluation on real-world data demonstrates that the framework improves both predictive performance and model traceability. 
These results demonstrate that linguistically grounded modifications to transformer self-attention improve the detection of communicative meaning, generalize beyond benchmark datasets to real-world mental-health data, and yield traceable, role-grounded explanations that extend beyond predictive performance to interpretability.

\end{abstract}


\section{Introduction}


Linguistic signals of internal states (e.g., sentiment, emotion) and external cues (e.g., politeness, formality) are often expressed through relational interaction patterns. Accurately understanding communication requires modeling these structures. Prior work in psychology, sociolinguistics, and communication theories examines these phenomena through constructs such as affective valence~\cite{russell1980circumplex}, basic emotion categories~\cite{ekman1971constants}, politeness strategies~\cite{brown1978universals}, and interpersonal formality~\cite{heylighen1999formality}. These theories characterize participant roles as central to understanding these dimensions; however, computational models of these dimensions largely do not.

\begin{figure}[h!]
\centering
\includegraphics[width=.85\linewidth]{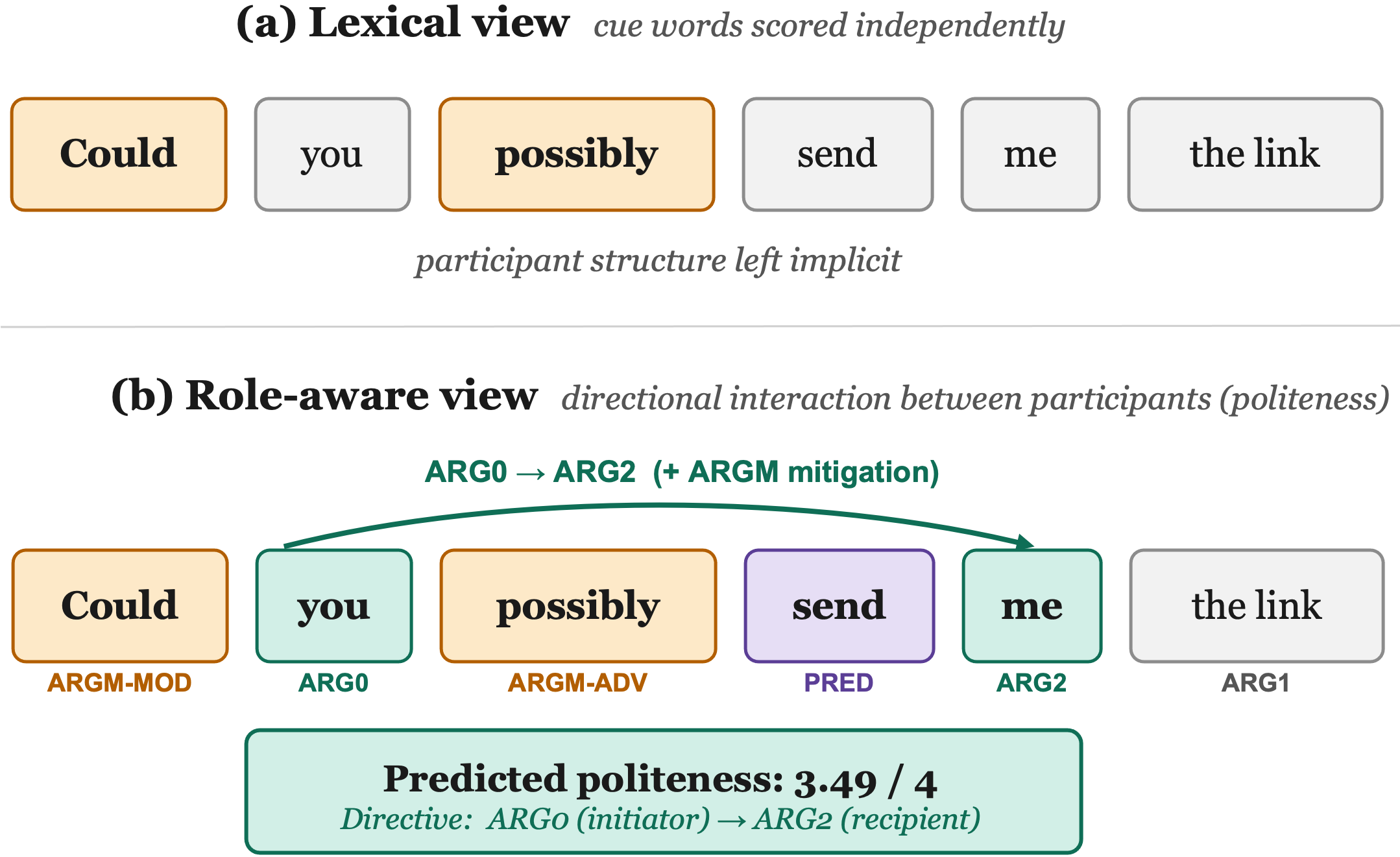}
\caption{Lexical and role-aware are compared using a politeness example. Lexical approaches score cue words (e.g., ``Could,'' ``possibly''), leaving participant structure implicit. \titleac{} models the 
interaction from initiator (ARG0) to recipient (ARG2), with mitigation modifiers (ARGM) softening the directive. Here, a politeness score (3.49) is predicted.}

\label{fig:intro_example}
\end{figure}

In NLP, these tasks are typically addressed by fine-tuning models using lexical features alone, treating sentiment, emotion, politeness, and formality (SEPF) as separate surface-level classification problems.
(See Figure~\ref{fig:intro_example}(a).) When additional features are incorporated, they are typically added through auxiliary concatenation without theory-grounded principles guiding which semantic-role information to incorporate, motivating the question of whether explicitly modeling participant roles improves detection of \textit{internal states} (sentiment, emotion) and \textit{external cues} (politeness, formality). These two groups differ fundamentally: internal states describe properties of a single participant, whereas external cues arise from relationships between participants.


Semantic role labeling (SRL) captures participant relations through structured representations of \textit{who did what to whom}. These representations support downstream NLP tasks such as question answering~\cite{he-etal-2015-question}, summarization~\cite{genest-lapalme-2011-framework}, cross-lingual SRL projection~\cite{youm2024dahrs}, and stance detection~\cite{mather2021general}. In Figure~\ref{fig:intro_example}(b), SRL assigns ARG0 to the initiator (``you''), ARG1 to the object being sent (``the link''), and ARG2 to the recipient (``me'') in the utterance ``Could you possibly send me the link?'' This structure supports role-based reasoning, e.g., ``Who is being asked to send the link?'' is answered by identifying ARG0 as \textit{you}.

Building on this role-aware perspective, \titleac{} represents each utterance along three relational dimensions: who is speaking, what is being discussed, and who is being addressed. In Figure~\ref{fig:intro_example}, ``you'' is the speaker, ``the link'' is the subject of the request, and ``me'' is the recipient. \titleac{} incorporates the corresponding role configurations into transformer self-attention and fuses the resulting role-aware and contextual representations through a learned gate. 

This architecture differs from SemBERT~\cite{zhang2020semantics}, which encodes semantic-role labels as embeddings and concatenates them with BERT-derived contextual word representations. In contrast, \titleac{} injects role information directly into a role-aware self-attention mechanism that determines which roles are most relevant to each prediction. Because modern SRL models automatically assign semantic roles, \titleac{} requires no manual role annotation, enabling scalable enrichment of existing datasets.

The contributions of this work are as follows:

\begin{enumerate}
\item \textbf{Theory-grounded role-aware framework:} A linguistically grounded mapping that associates sentiment, emotion, politeness, and formality (SEPF) with semantic-role configurations derived from established communication theories, treating SRL as a theory-driven representation rather than a generic feature set.
\item \textbf{Role-aware transformer architecture:} Complementary role-aware and directional role-aware attention mechanisms that integrate semantic-role configurations directly into transformer self-attention and fuse them with contextual representations through a learned gate, allowing role information to influence contextualization rather than appending it as an auxiliary feature.
\item \textbf{Benchmark and real-world evaluation:} A comprehensive evaluation across benchmark datasets together with an applicability study on real-world mental-health dialogue, demonstrating improved detection of communicative meaning and yielding traceable, role-grounded explanations without sacrificing predictive performance.
\end{enumerate}


Across four benchmark datasets, \titleac{} outperforms strong BERT-based baselines for emotion, politeness, and formality, with 
largest gains in theoretically predicted cases such as low-frequency emotions and highly polite expressions. The framework 
incorporates interpretable semantic-role representations without sacrificing predictive performance, improving on transformer baselines across 
settings.

\section{Related Work}


We review computational approaches to sentiment, emotion, politeness, and formality detection. Although these phenomena are typically modeled as separate tasks, \titleac{} treats these phenomena as expressions of a shared interactional structure involving participants, events, and evaluative context. Within this framework, sentiment and emotion are modeled as internal states, whereas politeness and formality are modeled as external cues. This perspective motivates the theory-grounded mappings presented in the next section.

\textbf{Sentiment Detection.} 
Sentiment detection methods typically fall into four categories~\cite{mao2024sentiment}. Lexicon-based approaches assign sentiment scores to individual tokens~\cite{mitra2021sentiment} but often require domain-specific lexicons or annotations. Traditional machine learning models~\cite{rahat2019comparison, bayhaqy2018sentiment} learn from engineered features and often outperform lexicon-based methods. BERT-based models~\cite{joshy2022analyzing, kewalramani2024optimization} use bidirectional contextual representations and typically achieve higher accuracy~\cite{mao2024sentiment}. Decoder-only LLMs perform sentiment detection through prompting~\cite{zhang2024sentiment} and can be effective in low-resource settings but may struggle with phenomena requiring deeper affective interpretation. \titleac{} models sentiment through a theory-grounded mapping in which semantic-role configurations link events to their evaluative context.

\textbf{Emotion Detection.} Emotion detection is typically formulated as a classification task.
Early studies rely on supervised classifiers~\cite{chaffar2011using, sailunaz2019emotion} with bags-of-words, n-grams, or part-of-speech features to detect Ekman's six emotions~\cite{ekman1971constants}. Embedding-based neural models~\cite{kratzwald2018deep, 10.1007/s11042-019-07813-9} have transcended traditional classifiers
but remain limited to coarse emotion categories.
BERT-family models~\cite{devlin-etal-2019-bert} 
support classification under Plutchik's eight-emotion framework~\cite{ruiz2025textual}, while GoEmotions~\cite{demszky-etal-2020-goemotions} introduces 27 fine-grained categories. Recent work explores prompt-based emotion detection with LLMs~\cite{elyoseph2024capacity}. 
\titleac{} models emotional meaning through a theory-grounded mapping in which semantic-role configurations capture relationships among events, experiencers, and modifiers.

\textbf{Politeness Detection.} Politeness detection has progressed from feature-based classifiers to neural and transformer architectures. Early studies rely on SVM and CNN-based models with lexical and linguistic features~\cite{danescu-niculescu-mizil-etal-2013-computational, aubakirova-bansal-2016-interpreting}, followed by regression-based models~\cite{chhaya-etal-2018-frustrated} and graph-based convolutional networks jointly detecting politeness and emotion~\cite{10.1016/j.eswa.2023.120025}. Later transformer approaches include hierarchical models~\cite{mishra2022predicting}, BERT-based models~\cite{bharti2024politepeer}, and graph-induced architectures for cross-domain politeness detection~\cite{10.1145/3543873.3587352}. A RoBERTa model incorporating PoliteLex~\cite{li2020studying} improves interpretability by highlighting lexical and phrasal cues that best predict politeness predictions~\cite{havaldar-etal-2023-comparing}. \titleac{} extends lexical and phrasal cues, adopting a theory-grounded mapping in which politeness is represented by directional semantic-role configurations linking speakers, recipients, and mitigation strategies.

\textbf{Formality Detection.} Formality detection is typically formulated as a binary classification task. Linguistic features such as contractions, voice, and average word length have been used as input to traditional classifiers~\cite{sheikha2010automatic}. Subsequent work has combined FastText vectors~\cite{bojanowski-etal-2017-enriching} with linguistic features under logistic regression; applied Universal Sentence Encoder~\cite{yang-etal-2020-multilingual} with a CNN; and paired ELMo~\cite{peters-etal-2018-deep} embeddings with a BiLSTM~\cite{dementieva-etal-2023-detecting}. BERT and its variants, including RoBERTa~\cite{liu1907roberta} and DeBERTa~\cite{he2020deberta}, have subsequently been widely adopted~\cite{dementieva-etal-2023-detecting}, while decoder-only models have been explored for generative classification~\cite{radford2019language}. Among these approaches, a hybrid DeBERTa and BiLSTM model achieves the strongest reported performance. 
\titleac{} adopts a theory-grounded mapping in which structured participant relationships represent degrees of formality. This formulation predicts a continuous score from 0 to $+1$ rather than a binary label.

\section{Theory-Grounded Role Mappings}

While prior models treat sentiment, emotion, politeness, and formality (SEPF) as independent, surface-level tasks, none captures the shared interactional structure underlying them. We develop four theory-grounded role mappings that operationalize these communication constructs through interpretable semantic-role configurations. These mappings naturally divide into two classes: sentiment and emotion represent internal states centered on an individual participant, whereas politeness and formality represent external cues arising from relationships among participants. This distinction motivates the complementary attention mechanisms introduced in the following section while connecting affective-science and social-communication theories to structured relational representations suitable for transformer models.


Building upon affective valence theory~\cite{russell1980circumplex} and prior work linking evaluative meaning to linguistic expression, \titleac{} treats sentiment as a relational phenomenon grounded in event-centered evaluation. Under this framework, affective experience is defined by a core valence dimension, which manifests in language not as isolated emotional tokens but as an evaluative cognitive stance contextualizing specific semantic entities and situations. \titleac{} 
adopts the following theory-grounded mapping:

\begin{quote}
\textbf{Theory-Grounded Sentiment Mapping.} 
Sentiment is expressed through the interaction between an event and its evaluative contextualization, operationalized through the semantic-role configuration: Predicate (PRED) $+$ Evaluative Modifier (ARGM)
\end{quote}

where the predicate anchors the event being evaluated and modifier structures encode evaluative orientation, intensity, or contextual stance.

Building upon dimensional and categorical theories of emotion~\cite{russell1980circumplex, plutchik1997circumplex}, \titleac{} treats emotional meaning as relationally structured through interactions among experiencers, affective events, and contextual intensifiers. Emotional experience is not a flat psychological response but a multidimensional construct characterized by varying degrees of affective intensity and contextual triggers that link an experiencing subject to a specific event. \titleac{} 
adopts the following theory-grounded mapping for emotion:

\begin{quote}
\textbf{Theory-Grounded Emotion Mapping.} 
Emotion is expressed through the interaction among an affective event, the affected entity, and contextual markers of intensity or manner, operationalized through the semantic-role configuration: Predicate (PRED) $+$ Affected Entity (ARG1) $+$ Modifier (ARGM)
\end{quote}

where the predicate denotes the affective event, ARG1 identifies the affected entity, and modifier structures encode intensity, manner, or contextual amplification. 

Building upon politeness theory and face-management frameworks~\cite{brown1978universals}, \titleac{} treats politeness as a directional interactional phenomenon shaped by social hierarchy, mitigation, and interpersonal negotiation. Here, politeness manifests as an interpersonal strategy minimizing face-threatening acts, fundamentally relational and dependent on the social distance and power dynamics between a speaker and a hearer. \titleac{} 
adopts the following theory-grounded mapping:
\begin{quote}
\textbf{Theory-Grounded Politeness Mapping.} 
Politeness is expressed through directional interactions between communicative participants and linguistic mitigation strategies, operationalized through the semantic-role configuration: $\text{Initiator (ARG0)} \rightarrow \text{Recipient (ARG2)} + \text{Mitigation (ARGM)}$
\end{quote}

where ARG0 represents the initiator of the communicative act, ARG2 the recipient affected by the act, and modifier structures encode hedging, softening, or mitigation, tracking an interactional between two distinct communicants unlike configurations modeling static internal states.


Building upon sociolinguistic theories of formality and social distance~\cite{brown1978universals, heylighen1999formality}, \titleac{} treats formality as an interactional phenomenon reflected through explicit participant structuring and communicative distance. Interactional formality does not reside in isolated lexical choices but rather reflects the social distance, power dynamics, and relational constraints negotiated between conversational participants. 
\titleac{} 
adopts the following theory-grounded mapping:

\begin{quote}
\textbf{Theory-Grounded Formality Mapping.} 
Formality is expressed through structured participant relationships within communicative interactions, operationalized through the semantic-role configuration: $\text{Initiator (ARG0)} \rightarrow \text{Affected Entity (ARG1)}$
\end{quote}

where ARG0 represents the initiating participant and ARG1 the explicitly realized affected entity whose presence, omission, or structural positioning signals degrees of social distance and communicative formality.

\begin{figure*}[h!]
\centering
\scriptsize
\includegraphics[width=.75\linewidth]{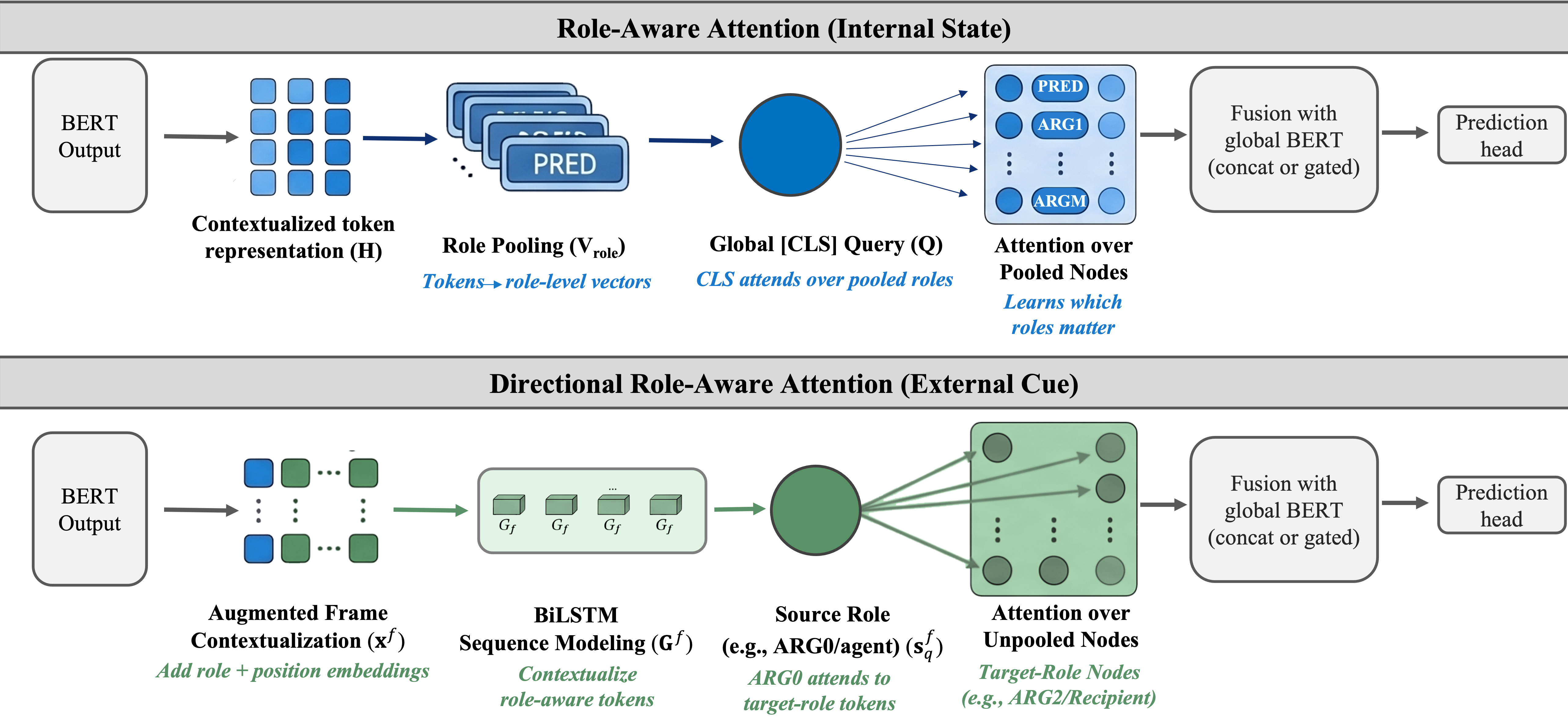}
\caption{\titleac{} uses complementary attention mechanisms for internal-state detection (top) and external-cue detection (bottom). For internal states, the global \texttt{[CLS]} query attends over pooled role vectors defined by the theory-grounded role mapping. For external cues, a source-role query (ARG0) attends over target-role positions to encode directional interactions between participants. Both branches fuse the role-aware representation with the BERT representation for prediction.} 
\label{fig:Chapter3_method}
\end{figure*}

\section{\titleac{} Framework}

This section presents the architectural implementation of the role-aware detection models. \titleac{} implements two complementary attention mechanisms reflecting the distinction between internal states and external cues. Rather than treating semantic roles as auxiliary features, the model integrates semantic-role representations defined by the theory-grounded mappings directly into transformer attention to guide contextual encoding and prediction. Internal states (sentiment and emotion) are modeled using role-aware attention over task-specific semantic roles, whereas external cues (politeness and formality) are modeled using directional role-aware attention that explicitly represents interactions between participants.

\subsection{Benchmark Datasets}
\label{sec:benchmark_chap3}

We introduce the benchmark datasets used to fine-tune and evaluate the SRL-informed transformer models. For internal-state detection, \titleac{} utilizes the GoEmotions dataset~\cite{demszky-etal-2020-goemotions}, which includes 27 emotion categories and a neutral label. For sentiment, these fine-grained categories are consolidated into four coarse labels (positive, negative, neutral, and ambiguous), while for emotion, the original categories are retained and mapped onto a two-dimensional valence-intensity space adapted from~\cite{russell1980circumplex}, where valence captures pleasantness and intensity represents the strength of emotional activation. For external-cue detection, \titleac{} utilizes the Holistic Politeness dataset~\cite{havaldar-etal-2023-comparing} for politeness and the GYAFC dataset~\cite{rao-tetreault-2018-dear} for formality, modeled as a binary classification task.





\subsection{Architectural Overview of SRL-Informed Detection}


This section presents the SRL-informed detection architecture (see Figure~\ref{fig:Chapter3_method}), in which role-aware representations are integrated directly into transformer-based contextual encoding and attention operations. For internal-state detection, the architecture selects task-specific semantic roles, pools their representations, and applies role-aware attention to construct a role-aware semantic representation. 

For detecting external cues, the architecture applies directional role-aware attention over the structural role configurations defined by the theory-grounded mappings for politeness and formality, augmenting frame-level representations with role and position embeddings and contextualizing them with a BiLSTM before directional attention. Both branches integrate the resulting role-aware representation with the BERT representation through feature concatenation or gated fusion before passing it to a task-specific prediction head. We describe these attention, fusion, and prediction mechanisms in detail below.

\subsection{Role-Aware and Directional Attention Mechanisms} 
\label{sec:Task-specific_SRL}


\titleac{} operationalizes the task-specific role configurations using two attention variants, extracting SRL via a BERT-based SRL framework~\cite{Youm2026RevisitingSRL}. Role-aware representations derived from these configurations are incorporated directly into the transformer model, rather than treated as static symbolic features or post-hoc annotations. Role-aware attention (internal states) asks \emph{which semantic roles are most important for predicting the target internal state}, whereas directional role-aware attention (external cues) asks \emph{how participant roles interact to convey an external cue}.

\textbf{Role pooling for internal states.}
For each semantic frame $f$ and role $r\in\mathcal{R}$, where $\mathcal{R}$ is the task role set ($\mathcal{R}_{\text{sent}}=\{\text{PRED}, \text{ARGM}\}$; $\mathcal{R}_{\text{emo}}=\{\text{PRED}, \text{ARG1}, \text{ARGM}\}$), the framework aggregates token representations assigned to role $r$ using masked mean pooling:
\begin{equation}
\mathbf{v}^{r}
=
\frac{\sum_{n=1}^{N} m_n^{r}\mathbf{H}_n^{f}}
{\max\left(\sum_{n=1}^{N}m_n^{r},\, 1\right)}
\in\mathbb{R}^{H},
\label{eq:role_pool}
\end{equation}
where $\mathbf{H}^{f}\in\mathbb{R}^{N\times H}$ denotes contextualized representations for frame $f$ and $m_n^r$ indicates whether token $n$ is assigned to role $r$. Pooled vectors are stacked into $\mathbf{V}_{\mathrm{role}}\in\mathbb{R}^{|\mathcal{R}|\times H}$, and a role-existence indicator $e^{r}=\mathbb{1}\!\left[\sum_n m_n^{r}>0\right]$ prevents the attention layer from attending to unrealized roles.

\textbf{Role-aware attention.}
The pooled \texttt{[CLS]} representation $\mathbf{p}\in\mathbb{R}^{H}$ serves as a query over $\mathbf{V}_{\mathrm{role}}$ via learned projections 
\begin{equation}
\mathbf{q}
=
\mathbf{W}_{Q}\mathbf{p},
\qquad
\mathbf{K}
=
\mathbf{V}_{\mathrm{role}}
\mathbf{W}_{K}^{\top},
\qquad
\mathbf{U}
=
\mathbf{V}_{\mathrm{role}}
\mathbf{W}_{V}^{\top},
\end{equation}


\begin{equation}
\alpha_{r}
=
\mathrm{softmax}_{r\in\mathcal{R}}
\!\left(
\frac{\mathbf{q}^{\top}\mathbf{K}_{r}}{\sqrt{d_a}}
+ \log e^{r}
\right),
\qquad
\mathbf{c}^{f}
=
\sum_{r\in\mathcal{R}}
\alpha_{r}\mathbf{U}_{r},
\label{eq:role_ctx}
\end{equation}
where unrealized roles ($e^r{=}0$) receive $-\infty$ scores. A projection layer yields the frame-level role-aware feature $\mathbf{r}^{f}=\tanh(\mathrm{LayerNorm}(\mathbf{W}_{o}\mathbf{c}^{f}))$, and frame-level features are aggregated across valid semantic frames using masked mean pooling into the utterance-level role-aware semantic representation $\mathbf{r}$.

\textbf{Directional role-aware attention for external cues.}
Unlike internal-state detection, which pools roles to model their presence, external-cue detection preserves frame-level token structure to represent source--target role relations. For each frame, token representations are augmented with role-indicator and predicate-relative position embeddings, $\mathbf{x}^{f}_{n}=[\mathbf{H}^{f}_{n};\mathbf{e}^{\mathrm{role}}_{n};\mathbf{e}^{\mathrm{dist}}_{n}]$, and contextualized by a BiLSTM, $\mathbf{G}^{f}=\mathrm{BiLSTM}(\mathbf{x}^{f}_{1:N})$. The query is constructed from the source role (ARG0), obtained by mean-pooling its contextualized representations, $\mathbf{s}^{f}_{q}$, while keys and values are computed over token positions and masked to the target roles defined by the corresponding theory-grounded mapping:
\begin{equation}
\mathbf{Q}^{f}_{\mathrm{ARG0}}
\rightarrow
(\mathbf{K}^{f}_{\mathrm{ARG2}\cup\mathrm{ARGM}},\,
\mathbf{V}^{f}_{\mathrm{ARG2}\cup\mathrm{ARGM}})
\quad\text{(politeness)},
\end{equation}
\begin{equation}
\mathbf{Q}^{f}_{\mathrm{ARG0}}
\rightarrow
(\mathbf{K}^{f}_{\mathrm{ARG1}},\,
\mathbf{V}^{f}_{\mathrm{ARG1}})
\quad\text{(formality)}.
\end{equation}

Attention weights are computed analogously to Eq.~\eqref{eq:role_ctx}, with the softmax taken over attendable target positions $n\in\mathcal{T}$ (in place of the role-existence mask), and the directional context is $\mathbf{c}^{f}=\mathrm{LayerNorm}\!\left(\sum_{n\in\mathcal{T}}\alpha_n^f\mathbf{V}^{f}_{n}\right)$. Directionality is encoded by the source-role query and target-role mask, not by the projection matrices.

\subsection{Fusion and Prediction}
\label{sec:SRL Integration}

In \titleac{}, role-aware information complements rather than replaces the original BERT representation. BERT provides contextual representations, while SRL-informed operations transform them into task-specific structural features. The framework then combines the role-aware semantic representation $\mathbf{r}_c\in\mathbb{R}^{d_r}$ with the global BERT representation $\mathbf{p}\in\mathbb{R}^{H}$ before prediction, preserving contextual information while incorporating role-informed structure. We evaluate two fusion strategies.

\textbf{Feature concatenation} directly appends the two representations immediately before prediction:
\begin{equation}
\mathbf{h}_{\mathrm{concat}} = [\mathbf{p}; \mathbf{r}_c],
\label{eq:concat}
\end{equation}
preserving both representations in their original form but treating role-aware information identically for every input. For internal-state detection, $\mathbf{r}_c$ corresponds to the role-aware attention representation constructed from the designated role set for cue $c$; for external-cue detection, it corresponds to the directional role-aware representation constructed from the source-role query and attended target-role context.

\textbf{Gated fusion} instead learns how strongly the role-aware representation should update the global representation for each input. Because the role-aware and global representations may reside in different representation spaces, $\mathbf{r}_c$ is first aligned to the hidden space of $\mathbf{p}$:
\begin{equation}
\tilde{\mathbf{r}}_c
=
\mathbf{W}_{r}\mathbf{r}_c+\mathbf{b}_{r},
\qquad
\tilde{\mathbf{r}}_c\in\mathbb{R}^{H}.
\label{eq:gate_proj}
\end{equation}
An input-specific gate, computed from both representations, then determines how much of this aligned representation contributes as a gated residual update:
\begin{equation}
\mathbf{g}_c
=
\sigma\!\left(\mathbf{W}_{g}[\mathbf{p};\mathbf{r}_c]+\mathbf{b}_{g}\right),
\qquad
\mathbf{h}_{\mathrm{fused}}
=
\mathbf{p}+\mathbf{g}_c\odot\tilde{\mathbf{r}}_c,
\label{eq:gate_fuse}
\end{equation}
where $\sigma(\cdot)$ is the elementwise sigmoid function and $\odot$ denotes elementwise multiplication, allowing the gate $\mathbf{g}_c\in\mathbb{R}^{H}$ to modulate each hidden dimension independently. The residual formulation enables the global contextual representation to be retained when the gate suppresses role-aware information, such as when an utterance lacks an informative role structure. Role information can refine, rather than overwrite, the contextual encoding.

Feature concatenation asks the classifier to use both representations directly, whereas gated fusion gives the model control over how much role-aware information influences each individual prediction; comparing the two tests whether adaptive integration provides additional benefit over direct concatenation. The final representation $\mathbf{h}_c$ is passed to a task-specific prediction head, implemented as either a linear layer or an MLP with one nonlinear hidden layer.

\section{Experiments and Results}


We evaluate the SRL-informed BERT models against baseline BERT models without SRL-informed components, across the architectural configurations, using each benchmark dataset's official train/dev/test split for reproducibility; the development set is used for checkpoint selection, and final performance is reported on the held-out test set. Hyperparameters are selected via a grid search over learning rate $\{5\times10^{-5}, 2\times10^{-5}, 1\times10^{-5}\}$, batch size $\{16, 32\}$, and epochs $\{4, 10\}$. The best configuration (learning rate $5\times10^{-5}$, batch size 16, trained for at least four epochs) is optimized using AdamW on NVIDIA B200 GPUs, with a fixed random seed (42), and all results are reported from a single training run per configuration.



For each dimension, two baseline BERT models (without SRL attention; linear and MLP prediction heads) are compared against four SRL-informed configurations, combining feature concatenation or gated fusion with a linear or MLP prediction head, resulting in six total model configurations for comparison. Sentiment and emotion detection are evaluated using macro-$F_1$; politeness predicts a continuous score on a 0\textendash 4 scale, ranging from rude to polite, and is evaluated using Pearson's correlation coefficient ($r$), which measures the linear alignment between predicted scores and gold-standard politeness ratings; formality is evaluated as a binary classification task using $F_1$. Table~\ref{tab:main_results} reports overall performance across all four dimensions.

\begin{table}[t]
\small
\centering

\caption{SRL-informed architectures outperform baseline models across the four communication tasks. Sentiment and emotion report macro-$F_1$, politeness reports Pearson's $r$, and formality reports $F_1$. Best results are shown in bold. McNemar's tests indicate statistically significant improvements for emotion and formality, but not sentiment.}
\label{tab:main_results}
\resizebox{\linewidth}{!}{%
\begin{tabular}{lcccc}
\toprule
\textbf{Model} & \textbf{Sentiment} & \textbf{Emotion} & \textbf{Politeness} & \textbf{Formality} \\
 & Macro-$F_1$ & Macro-$F_1$ & Pearson's $r$ & $F_1$ \\
\midrule
\textit{Baseline Models} & & & & \\
\quad BERT (linear) & 0.6419 & 0.4594 & 0.6121 & 0.9342 \\
\quad BERT (MLP) & 0.6594 & 0.4373 & 0.6222 & 0.9225 \\
\midrule
\textit{SRL-informed Models} & & & & \\
\quad Gated fusion, linear & 0.6594 & \textbf{0.4803} & \textbf{0.6412} & \textbf{0.9480} \\
\quad Gated fusion, MLP & 0.6645 & 0.4478 & 0.6184 & 0.9420 \\
\quad Feature concatenation, linear & \textbf{0.6695} & 0.4610 & 0.6372 & 0.9378 \\
\quad Feature concatenation, MLP & 0.6599 & 0.4457 & 0.6340 & 0.9440 \\
\bottomrule
\end{tabular}
}
\end{table}


\textbf{Sentiment Detection.} Among all configurations, the SRL-informed BERT model utilizing feature concatenation with a linear prediction head achieves the strongest overall performance (macro-$F_1$: 0.6695). One possible explanation is that the task-specific SRL-informed attention representations are already sufficiently informative for sentiment classification, allowing the classifier to benefit from a more direct mapping between the learned representation and the final output labels. However, McNemar's test on paired exact-match correctness against the strongest baseline, BERT (MLP), does not show a statistically significant difference ($p = .1527$). While the SRL-informed model yields modest class-level $F_1$ improvements for negative (0.6695 vs. 0.6582) and neutral (0.5997 vs. 0.5610) sentiment, the overall macro-$F_1$ improvement remains marginal.

\textbf{Emotion Detection.} The SRL-informed BERT model with gated fusion and a linear prediction layer achieves the highest macro-$F_1$ score, and McNemar's test against the strongest baseline, BERT (linear), shows a statistically significant difference in paired classification correctness ($p = .0037$). This suggests that the gate's dynamic updating property helps prevent overfitting to dominant categories, preserving distinguishing features for low-frequency classes. At the class level, the SRL-informed model achieves higher $F_1$ scores across 18 emotion categories and the neutral category, with the performance advantage particularly evident in low-frequency, high-complexity classes such as remorse (0.651 vs. 0.636), fear (0.667 vs. 0.613), and embarrassment (0.421 vs. 0.386). For the rare categories grief, relief, and pride, $F_1$ improves from 0.000 to 0.286, 0.167 to 0.222, and 0.316 to 0.476, respectively, indicating that role-aware representations capture rare emotional expressions beyond surface-level contextual encoding.


\textbf{Politeness Detection.} The SRL-informed BERT model configured with gated fusion and a linear prediction head achieves the highest overall Pearson correlation coefficient. This result suggests that configurations utilizing directional, role-informed attention are better positioned to process instances where politeness signaling correlates with contextual argument dependencies, which are explicitly modeled through the structural constraints of core predicate-argument relationships ($\text{ARG0}$ through $\text{ARG2}$) and circumstantial modifiers (ARGM). A bucket-level analysis across politeness ranges shows that the directional SRL-informed model achieves its larger gains within the higher-politeness intervals, exhibiting a distinct positive delta in both the Slightly Polite ($r = 0.299$ vs. $0.246$; $+0.053$) and Polite ($r = 0.435$ vs. $0.326$; $+0.109$) buckets, while the performance margins favoring the baseline remain narrow across the middle-to-lower buckets; these higher-interval improvements drive the higher overall Pearson correlation, indicating superior global alignment across the entire politeness scale.

\textbf{Formality Detection.} Across all model configurations, the SRL-informed BERT model using gated fusion and a linear prediction head achieves the highest $F_1$ score (0.9480) and recall (0.9345). This improvement is confirmed by McNemar's test against the strongest baseline, BERT (linear), with the difference statistically significant well below the .01 level, further evidence that gated fusion improves formality detection ($p = 3.0 \times 10^{-5}$).

Across the four detection dimensions, gated fusion with a linear prediction head is the best-performing configuration on three tasks. It also yields the most consistent performance overall.

\section{Applicability Study: Real-World Mental-Health Data} Building on the role-aware detection of internal states and external cues under structural linguistic variation, this section extends the SRL-informed framework to real-world mental health data. In doing so, it provides evidence of applicability beyond benchmark evaluations by examining whether the role-aware framework remains effective and traceable on naturally occurring data.

\begin{table}[h]
\small
\footnotesize
\centering
\caption{Model performance on the EMA dataset. Feature concatenation yields the strongest performance. Pearson's $r$ is reported, with the best result in bold.}
\label{tab:ema_pearson_correlation_results}
\resizebox{\linewidth}{!}{%
\begin{tabular}{lcc}
\toprule
\textbf{Model} & \textbf{Pearson's $r$} & \textbf{\textit{p}-value} \\ 
\midrule
\textit{Baseline Models} & & \\
\quad BERT (linear) & 0.5687 & $3.37 \times 10^{-161}$ \\
\quad BERT (MLP) & 0.5529 & $1.16 \times 10^{-150}$ \\ \midrule
\textit{SRL-informed Models} & & \\
\quad Gated fusion, linear & 0.5254 & $1.29 \times 10^{-133}$ \\
\quad Gated fusion, MLP & 0.5526 & $1.75 \times 10^{-150}$ \\
\quad \textbf{Feature concatenation, linear} & \textbf{0.5754} & $7.05 \times 10^{-166}$ \\
\quad Feature concatenation, MLP & 0.5353 & $1.29 \times 10^{-133}$ \\ 
\bottomrule
\end{tabular}
}
\end{table}

We use an ecological momentary assessment (EMA) dataset collected 
via the EARS app provided by Ksana Health (\url{https://ksanahealth.com/ears/}) as part of a study on youth mental health. The sample consists of 74 adolescents aged 13--15, monitored over one or two 21-day bursts. 
Daily EMA surveys capture self-reported ratings of 14 emotions (e.g., happy, embarrassed, bored) on a 1--5 Likert scale, and the platform separately logs outgoing text messages; pairing these two streams yields 1,924 messages with corresponding emotion ratings. The study was IRB-approved and conducted with full participant consent. This setting provides a particularly demanding test of the role-aware framework because the language is informal, context-dependent, naturalistic, and captured during real-world mental-health monitoring rather than elicited under controlled experimental conditions.

\begin{table}[t]
\centering
\scriptsize
\caption{Post-hoc explanations generated with and without SRL grounding are compared using a politeness example from the four communication tasks. The \textit{No SRL} explanation identifies lexical and grammatical features. The \textit{With SRL} explanation refers to the same role components used by the model's role-aware attention pattern, illustrating that SRL-grounded explanations remain tied to \titleac{}'s role-aware components used during prediction.}
\label{tab:srl_explanation_examples}
\begin{tabular}{@{}p{0.96\linewidth}@{}}
\toprule
\textbf{Politeness} (range $0$ to $4$) \quad Utterance: ``Could you possibly send me the link?'' \quad Predicted: 3.49 \\
\midrule
\textbf{No SRL} \\
\textit{Input:} Utterance + score \\
\textbf{LLM Explanation:} The use of the modal verb ``Could'' combined with the conditional phrasing ``you possibly'' softens the request, and the interrogative form signals deference; these cues may be interpreted as indicating courteous intent. \\
\midrule
\textbf{With SRL} \\
\textit{SRL Input:} [ARGM-MOD: Could] [ARG0: you] [ARGM-ADV: possibly] [V: send] [ARG2: me] [ARG1: the link] \newline
\textit{Semantic-role repr.:} ARG0 $\rightarrow$ ARG2 $+$ ARGM \\
\textbf{LLM Explanation:} The utterance frames the agent (you) as the potential performer of the action (send) toward the recipient (me) with the object (the link). The presence of the modal ``Could'' and the adverb ``possibly'' softens the directive, which may be interpreted as a polite request, consistent with the high politeness score. \\
\bottomrule
\end{tabular}
\end{table}

\subsection{Role-Aware Emotion Detection on EMA Data}


Because the data include only participants' outgoing text messages, without recipients' replies, and ground-truth labels are available exclusively for self-reported emotion ratings, this evaluation focuses on emotion detection. It examines whether the SRL-informed model retains its advantage over baseline transformers when fine-tuned on EMA data. Before evaluating the fine-tuned models, a chi-square test of independence confirms that the existence of the core predicate-argument configuration ($\text{PRED} + \text{ARG1} + \text{ARGM}$) correlates with a higher baseline likelihood of target emotional expressions across all 14 evaluated emotion categories ($\chi^2 = 47.87$, $p = 6.87 \times 10^{-6}$).



As shown in Table~\ref{tab:ema_pearson_correlation_results}, incorporating semantic-role representations yields the strongest overall performance. The SRL-informed BERT model using feature concatenation with a linear prediction head achieves a Pearson correlation of 0.5754, exceeding the strongest baseline (0.5687). This demonstrates that semantic-role information provides complementary predictive value beyond contextualized transformer representations alone. At the emotion level, the SRL-informed model achieves statistically significant correlations ($p < 0.05$) for 9 of the 14 evaluated emotion categories, with gains over the baseline for tired (0.4807 vs. 0.3758), bored (0.2975 vs. 0.1965), and cheerful (0.2828 vs. 0.2474); the baseline performs better on four emotions (sad, irritable, accepted, lonely), while neither model reaches significance for happy.


\subsection{Interpretability of Role-Aware Detection}

Beyond predictive performance, post-hoc explanations provide human-readable rationales for model predictions~\cite{ribeiro-etal-2016-trust, lundberg2017unified}, but often suffer from a faithfulness problem, generating plausible explanations that may not accurately reflect the model's underlying reasoning~\cite{jacovi2020towards,camburu2019can}. Unlike explanation generators without access to internal representations, \titleac{} architecturally exposes its role-aware representations to the explanation generator, making the correspondence between explanations and these representations inspectable.


To examine whether semantic-role representations support verifiable explanations, two prompt configurations are tested: with and without SRL grounding. For each prediction, a zero-shot LLM (gpt-oss-120b) is prompted with the original utterance, its SRL-parsed predicate-argument structure, the semantic-role representation specific to each dimension, and the framework's predicted score. The baseline prompt omits the SRL-parsed structure and the semantic-role representation while keeping all other elements identical, isolating the contribution of SRL grounding.

Table~\ref{tab:srl_explanation_examples} extends the politeness example with a full explanation comparison. Here, the baseline identifies lexical cues (e.g., \textit{could}, \textit{possibly}) without characterizing participant structure, whereas the SRL-grounded explanation explicitly references ARG0 and ARG2 to describe the same directive-softening pattern. This qualitative analysis targets NLP researchers and system developers, providing a structural prerequisite for, though not equivalent to, full interpretability claims requiring systematic human evaluation. This additional interpretability is obtained without sacrificing predictive accuracy and, in several settings, yields measurable improvements over strong transformer baselines.

\section{Conclusion}

\titleac{} integrates semantic-role-conditioned attention mechanisms directly into transformer-based contextual representations for detecting internal states and external cues. We demonstrate measurable improvements across all four detection tasks relative to baseline transformer architectures, with the gains in emotion, politeness, and formality proving statistically significant. This architecture combines multiple integration strategies (role-aware attention, directional attention, feature concatenation, and gated fusion) to model predicate-argument dependencies across sentiment, emotion, politeness, and formality. This enables analysis of which role configurations are emphasized during prediction, making the semantic-role basis of each prediction interpretable.

Beyond benchmark datasets, \titleac{} extends to real-world mental-health data, improving emotion prediction over baseline transformers and making the semantic-role basis of each prediction interpretable. This interpretability analysis is qualitative and illustrative. Future work includes systematic, large-scale evaluation of structural correspondence, as well as perturbation-based tests of explanation faithfulness.




\bigskip
\bibliography{references_updates}

@inproceedings{zhang2020semantics,
  title={Semantics-aware BERT for language understanding},
  author={Zhang, Zhuosheng and Wu, Yuwei and Zhao, Hai and Li, Zuchao and Zhang, Shuailiang and Zhou, Xi and Zhou, Xiang},
  booktitle={Proceedings of the AAAI conference on artificial intelligence},
  volume={34},
  number={05},
  pages={9628--9635},
  year={2020}
}

@inproceedings{youm2024dahrs,
  title={DAHRS: Divergence-Aware Hallucination-Remediated SRL Projection},
  author={Youm, Sangpil and Mather, Brodie and Jayaweera, Chathuri and Prada, Juliana and Dorr, Bonnie},
  booktitle={International Conference on Applications of Natural Language to Information Systems},
  pages={423--438},
  year={2024},
  organization={Springer},
  addendum={\url{https://link.springer.com/chapter/10.1007/978-3-031-70239-6_29}}
}

@article{liu1907roberta,
  title={RoBERTa: A robustly optimized BERT pretraining approach. arXiv [Preprint](2019)},
  author={Liu, Y and Ott, M and Goyal, N and Du, J and Joshi, M and Chen, D and Levy, O and Lewis, M and Zettlemoyer, L and Stoyanov, V},
  journal={arXiv preprint arXiv:1907.11692},
  year={1907},
 addendum={\url{https://doi.org/10.48550/arXiv.1907.11692}}
}

@article{radford2019language,
  title={Language models are unsupervised multitask learners},
  author={Radford, Alec and Wu, Jeffrey and Child, Rewon and Luan, David and Amodei, Dario and Sutskever, Ilya and others},
  journal={OpenAI blog},
  volume={1},
  number={8},
  pages={9},
  year={2019}
}

@inproceedings{mather2021general,
  title={A general framework for domain-specialization of stance detection: a COVID-19 response use case},
  author={Mather, Brodie and Dorr, Bonnie J and Rambow, Owen and Strzalkowski, Tomek},
  booktitle={The International FLAIRS Conference Proceedings},
  volume={34},
  year={2021}
}

@inproceedings{genest-lapalme-2011-framework,
    title = "Framework for Abstractive Summarization using Text-to-Text Generation",
    author = "Genest, Pierre-Etienne  and
      Lapalme, Guy",
    editor = "Filippova, Katja  and
      Wan, Stephen",
    booktitle = "Proceedings of the Workshop on Monolingual Text-To-Text Generation",
    month = jun,
    year = "2011",
    address = "Portland, Oregon",
    publisher = "Association for Computational Linguistics",
    url = "https://aclanthology.org/W11-1608",
    pages = "64--73",
}

@inproceedings{havaldar-etal-2023-comparing,
    title = "Comparing Styles across Languages",
    author = "Havaldar, Shreya  and
      Pressimone, Matthew  and
      Wong, Eric  and
      Ungar, Lyle",
    editor = "Bouamor, Houda  and
      Pino, Juan  and
      Bali, Kalika",
    booktitle = "Proceedings of the 2023 Conference on Empirical Methods in Natural Language Processing",
    month = dec,
    year = "2023",
    address = "Singapore",
    publisher = "Association for Computational Linguistics",
    url = "https://aclanthology.org/2023.emnlp-main.419/",
    doi = "10.18653/v1/2023.emnlp-main.419",
    pages = "6775--6791"
}

@inproceedings{rao-tetreault-2018-dear,
    title = "Dear Sir or Madam, May {I} Introduce the {GYAFC} Dataset: Corpus, Benchmarks and Metrics for Formality Style Transfer",
    author = "Rao, Sudha  and
      Tetreault, Joel",
    editor = "Walker, Marilyn  and
      Ji, Heng  and
      Stent, Amanda",
    booktitle = "Proceedings of the 2018 Conference of the North {A}merican Chapter of the Association for Computational Linguistics: Human Language Technologies, Volume 1 (Long Papers)",
    month = jun,
    year = "2018",
    address = "New Orleans, Louisiana",
    publisher = "Association for Computational Linguistics",
    url = "https://aclanthology.org/N18-1012/",
    doi = "10.18653/v1/N18-1012",
    pages = "129--140"
}

@incollection{brown1978universals,
  title={Universals in language usage: Politeness phenomena},
  author={Brown, Penelope and Levinson, Stephen C.},
  booktitle={Questions and politeness: Strategies in social interaction},
  pages={56--311},
  year={1978},
  publisher={Cambridge University Press}
}

@article{heylighen1999formality,
  title={Formality of language: definition, measurement and behavioral determinants},
  author={Heylighen, Francis and Dewaele, Jean-Marc},
  journal={Interner Bericht, Center “Leo Apostel”, Vrije Universiteit Br{\"u}ssel},
  volume={4},
  number={1},
  year={1999},
  publisher={Citeseer}
}

@inproceedings{demszky-etal-2020-goemotions,
    title = "{G}o{E}motions: A Dataset of Fine-Grained Emotions",
    author = "Demszky, Dorottya  and
      Movshovitz-Attias, Dana  and
      Ko, Jeongwoo  and
      Cowen, Alan  and
      Nemade, Gaurav  and
      Ravi, Sujith",
    booktitle = "Proceedings of the 58th Annual Meeting of the Association for Computational Linguistics",
    month = jul,
    year = "2020",
    url = "https://aclanthology.org/2020.acl-main.372/",
    doi = "10.18653/v1/2020.acl-main.372",
    pages = "4040--4054"
}

@inproceedings{danescu-niculescu-mizil-etal-2013-computational,
    title = "A computational approach to politeness with application to social factors",
    author = "Danescu-Niculescu-Mizil, Cristian  and
      Sudhof, Moritz  and
      Jurafsky, Dan  and
      Leskovec, Jure  and
      Potts, Christopher",
    editor = "Schuetze, Hinrich  and
      Fung, Pascale  and
      Poesio, Massimo",
    booktitle = "Proceedings of the 51st Annual Meeting of the Association for Computational Linguistics (Volume 1: Long Papers)",
    month = aug,
    year = "2013",
    address = "Sofia, Bulgaria",
    publisher = "Association for Computational Linguistics",
    url = "https://aclanthology.org/P13-1025/",
    pages = "250--259"
}

@book{plutchik1997circumplex,
  title={Circumplex models of personality and emotions.},
  author={Plutchik, Robert Ed and Conte, Hope R},
  year={1997},
  publisher={American Psychological Association}
}

@article{russell1980circumplex,
  title={A circumplex model of affect.},
  author={Russell, James A},
  journal={Journal of personality and social psychology},
  volume={39},
  number={6},
  pages={1161},
  year={1980},
  publisher={American Psychological Association}
}

@article{ekman1971constants,
  title={Constants across cultures in the face and emotion.},
  author={Ekman, Paul and Friesen, Wallace V},
  journal={Journal of personality and social psychology},
  volume={17},
  number={2},
  pages={124},
  year={1971},
  publisher={American Psychological Association}
}

@inproceedings{chaffar2011using,
  title={Using a heterogeneous dataset for emotion analysis in text},
  author={Chaffar, Soumaya and Inkpen, Diana},
  booktitle={Canadian conference on artificial intelligence},
  pages={62--67},
  year={2011},
  organization={Springer}
}

@inproceedings{peters-etal-2018-deep,
    title = "Deep Contextualized Word Representations",
    author = "Peters, Matthew E.  and
      Neumann, Mark  and
      Iyyer, Mohit  and
      Gardner, Matt  and
      Clark, Christopher  and
      Lee, Kenton  and
      Zettlemoyer, Luke",
    editor = "Walker, Marilyn  and
      Ji, Heng  and
      Stent, Amanda",
    booktitle = "Proceedings of the 2018 Conference of the North {A}merican Chapter of the Association for Computational Linguistics: Human Language Technologies, Volume 1 (Long Papers)",
    month = jun,
    year = "2018",
    address = "New Orleans, Louisiana",
    publisher = "Association for Computational Linguistics",
    url = "https://aclanthology.org/N18-1202/",
    doi = "10.18653/v1/N18-1202",
    pages = "2227--2237"
}

@inproceedings{devlin-etal-2019-bert,
    title = "{BERT}: Pre-training of Deep Bidirectional Transformers for Language Understanding",
    author = "Devlin, Jacob  and
      Chang, Ming-Wei  and
      Lee, Kenton  and
      Toutanova, Kristina",
    editor = "Burstein, Jill  and
      Doran, Christy  and
      Solorio, Thamar",
    booktitle = "Proceedings of the 2019 Conference of the North {A}merican Chapter of the Association for Computational Linguistics: Human Language Technologies, Volume 1 (Long and Short Papers)",
    month = jun,
    year = "2019",
    url = "https://aclanthology.org/N19-1423/",
    doi = "10.18653/v1/N19-1423",
    pages = "4171--4186",
  
}

@inproceedings{sheikha2010automatic,
  title={Automatic classification of documents by formality},
  author={Sheikha, Fadi Abu and Inkpen, Diana},
  booktitle={Proceedings of the 6th international conference on natural language processing and knowledge engineering (nlpke-2010)},
  pages={1--5},
  year={2010},
  organization={IEEE}
}

@inproceedings{dementieva-etal-2023-detecting,
    title = "Detecting Text Formality: A Study of Text Classification Approaches",
    author = "Dementieva, Daryna  and
      Babakov, Nikolay  and
      Panchenko, Alexander",
    editor = "Mitkov, Ruslan  and
      Angelova, Galia",
    booktitle = "Proceedings of the 14th International Conference on Recent Advances in Natural Language Processing",
    month = sep,
    year = "2023",
    address = "Varna, Bulgaria",
    publisher = "INCOMA Ltd., Shoumen, Bulgaria",
    url = "https://aclanthology.org/2023.ranlp-1.31/",
    pages = "274--284",
}

@article{li2020studying,
  title={Studying politeness across cultures using English Twitter and Mandarin Weibo},
  author={Li, Mingyang and Hickman, Louis and Tay, Louis and Ungar, Lyle and Guntuku, Sharath Chandra},
  journal={Proceedings of the ACM on human-computer interaction},
  volume={4},
  number={CSCW2},
  pages={1--15},
  year={2020},
  publisher={ACM New York, NY, USA}
}

@inproceedings{lundberg2017unified,
author = {Lundberg, Scott M. and Lee, Su-In},
title = {A unified approach to interpreting model predictions},
year = {2017},
isbn = {9781510860964},
publisher = {Curran Associates Inc.},
address = {Red Hook, NY, USA},
booktitle = {Proceedings of the 31st International Conference on Neural Information Processing Systems},
pages = {4768–4777},
numpages = {10},
location = {Long Beach, California, USA},
series = {NIPS'17}
}

@article{10.1016/j.eswa.2023.120025,
author = {Priya, Priyanshu and Firdaus, Mauajama and Ekbal, Asif},
title = {A multi-task learning framework for politeness and emotion detection in dialogues for mental health counselling and legal aid},
year = {2023},
issue_date = {Aug 2023},
publisher = {Pergamon Press, Inc.},
address = {USA},
volume = {224},
number = {C},
issn = {0957-4174},
url = {https://doi.org/10.1016/j.eswa.2023.120025},
doi = {10.1016/j.eswa.2023.120025},
journal = {Expert Syst. Appl.},
month = aug,
numpages = {17}
}

@article{sailunaz2019emotion,
  title={Emotion and sentiment analysis from Twitter text},
  author={Sailunaz, Kashfia and Alhajj, Reda},
  journal={Journal of computational science},
  volume={36},
  pages={101003},
  year={2019},
  publisher={Elsevier}
}

@article{10.1007/s11042-019-07813-9,
author = {Shrivastava, Kush and Kumar, Shishir and Jain, Deepak Kumar},
title = {An effective approach for emotion detection in multimedia text data using sequence based convolutional neural network},
year = {2019},
issue_date = {Oct 2019},
publisher = {Kluwer Academic Publishers},
address = {USA},
volume = {78},
number = {20},
issn = {1380-7501},
url = {https://doi.org/10.1007/s11042-019-07813-9},
doi = {10.1007/s11042-019-07813-9},
journal = {Multimedia Tools Appl.},
month = oct,
pages = {29607–29639},
numpages = {33}
}

@article{kratzwald2018deep,
  title={Deep learning for affective computing: Text-based emotion recognition in decision support},
  author={Kratzwald, Bernhard and Ili{\'c}, Suzana and Kraus, Mathias and Feuerriegel, Stefan and Prendinger, Helmut},
  journal={Decision support systems},
  volume={115},
  pages={24--35},
  year={2018},
  publisher={Elsevier}
}

@article{ruiz2025textual,
  title={Textual Emotion Detection with complementary BERT transformers in a Condorcet’s Jury Theorem assembly},
  author={Ruiz, Gerardo B{\'a}rcena and Herrera, Richard De Jes{\'u}s Gil},
  journal={Knowledge-Based Systems},
  pages={114070},
  year={2025},
  publisher={Elsevier}
}

@article{elyoseph2024capacity,
  title={Capacity of generative AI to interpret human emotions from visual and textual data: pilot evaluation study},
  author={Elyoseph, Zohar and Refoua, Elad and Asraf, Kfir and Lvovsky, Maya and Shimoni, Yoav and Hadar-Shoval, Dorit},
  journal={JMIR Mental Health},
  volume={11},
  pages={e54369},
  year={2024},
  publisher={JMIR Publications Toronto, Canada}
}

@article{bojanowski-etal-2017-enriching,
    title = "Enriching Word Vectors with Subword Information",
    author = "Bojanowski, Piotr  and
      Grave, Edouard  and
      Joulin, Armand  and
      Mikolov, Tomas",
    editor = "Lee, Lillian  and
      Johnson, Mark  and
      Toutanova, Kristina",
    journal = "Transactions of the Association for Computational Linguistics",
    volume = "5",
    year = "2017",
    address = "Cambridge, MA",
    publisher = "MIT Press",
    url = "https://aclanthology.org/Q17-1010/",
    doi = "10.1162/tacl_a_00051",
    pages = "135--146"
}

@inproceedings{yang-etal-2020-multilingual,
    title = "Multilingual Universal Sentence Encoder for Semantic Retrieval",
    author = "Yang, Yinfei  and
      Cer, Daniel  and
      Ahmad, Amin  and
      Guo, Mandy  and
      Law, Jax  and
      Constant, Noah  and
      Hernandez Abrego, Gustavo  and
      Yuan, Steve  and
      Tar, Chris  and
      Sung, Yun-hsuan  and
      Strope, Brian  and
      Kurzweil, Ray",
    editor = "Celikyilmaz, Asli  and
      Wen, Tsung-Hsien",
    booktitle = "Proceedings of the 58th Annual Meeting of the Association for Computational Linguistics: System Demonstrations",
    month = jul,
    year = "2020",
    address = "Online",
    publisher = "Association for Computational Linguistics",
    url = "https://aclanthology.org/2020.acl-demos.12/",
    doi = "10.18653/v1/2020.acl-demos.12",
    pages = "87--94"
}

@article{he2020deberta,
  title={Deberta: Decoding-enhanced bert with disentangled attention},
  author={He, Pengcheng and Liu, Xiaodong and Gao, Jianfeng and Chen, Weizhu},
  journal={arXiv preprint arXiv:2006.03654},
  year={2020}
}

@inproceedings{aubakirova-bansal-2016-interpreting,
    title = "Interpreting Neural Networks to Improve Politeness Comprehension",
    author = "Aubakirova, Malika  and
      Bansal, Mohit",
    editor = "Su, Jian  and
      Duh, Kevin  and
      Carreras, Xavier",
    booktitle = "Proceedings of the 2016 Conference on Empirical Methods in Natural Language Processing",
    month = nov,
    year = "2016",
    address = "Austin, Texas",
    publisher = "Association for Computational Linguistics",
    url = "https://aclanthology.org/D16-1216/",
    doi = "10.18653/v1/D16-1216",
    pages = "2035--2041"
}

@inproceedings{chhaya-etal-2018-frustrated,
    title = "Frustrated, Polite, or Formal: Quantifying Feelings and Tone in Email",
    author = "Chhaya, Niyati  and
      Chawla, Kushal  and
      Goyal, Tanya  and
      Chanda, Projjal  and
      Singh, Jaya",
    editor = "Nissim, Malvina  and
      Patti, Viviana  and
      Plank, Barbara  and
      Wagner, Claudia",
    booktitle = "Proceedings of the Second Workshop on Computational Modeling of People{'}s Opinions, Personality, and Emotions in Social Media",
    month = jun,
    year = "2018",
    address = "New Orleans, Louisiana, USA",
    publisher = "Association for Computational Linguistics",
    url = "https://aclanthology.org/W18-1111/",
    doi = "10.18653/v1/W18-1111",
    pages = "76--86",
}

@article{mishra2022predicting,
  author={Mishra, Kshitij and Firdaus, Mauajama and Ekbal, Asif},
  journal={IEEE Transactions on Computational Social Systems}, 
  title={Predicting Politeness Variations in Goal-Oriented Conversations}, 
  year={2023},
  volume={10},
  number={3},
  pages={1095-1104},
  doi={10.1109/TCSS.2022.3156580}}

@article{bharti2024politepeer,
        title={PolitePEER: does peer review hurt? A dataset to gauge politeness intensity in the peer reviews},
        author={Bharti, Prabhat Kumar and Navlakha, Meith and Agarwal, Mayank and Ekbal, Asif},
        journal={Language Resources and Evaluation},
        volume={58},
        number={4},
        pages={1291--1313},
        year={2024},
        publisher={Springer}}

@inproceedings{10.1145/3543873.3587352,
author = {Dasgupta, Tirthankar and Sinha, Manjira and Geetha Praveen, Chundru},
title = {Graph Induced Transformer Network for Detection of Politeness and Formality in Text},
year = {2023},
isbn = {9781450394192},
publisher = {Association for Computing Machinery},
address = {New York, NY, USA},
url = {https://doi.org/10.1145/3543873.3587352},
doi = {10.1145/3543873.3587352},
booktitle = {Companion Proceedings of the ACM Web Conference 2023},
pages = {221–224},
numpages = {4},
location = {Austin, TX, USA},
series = {WWW '23 Companion}
}

@article{mao2024sentiment,
  title={Sentiment analysis methods, applications, and challenges: A systematic literature review},
  author={Mao, Yanying and Liu, Qun and Zhang, Yu},
  journal={Journal of King Saud University-Computer and Information Sciences},
  volume={36},
  number={4},
  pages={102048},
  year={2024},
  publisher={Elsevier}
}

@incollection{mitra2021sentiment,
  title={Sentiment analysis using machine learning approaches},
  author={Mitra, Ayushi and Mohanty, Sanjukta},
  booktitle={Emerging Technologies in Data Mining and Information Security: Proceedings of IEMIS 2020, Volume 2},
  pages={63--68},
  year={2021},
  publisher={Springer}
}

@inproceedings{rahat2019comparison,
  title={Comparison of Naive Bayes and SVM Algorithm based on sentiment analysis using review dataset},
  author={Rahat, Abdul Mohaimin and Kahir, Abdul and Masum, Abu Kaisar Mohammad},
  booktitle={2019 8th International conference system modeling and advancement in research trends (SMART)},
  pages={266--270},
  year={2019},
  organization={IEEE}
}

@inproceedings{bayhaqy2018sentiment,
  title={Sentiment analysis about E-commerce from tweets using decision tree, K-nearest neighbor, and na{\"\i}ve bayes},
  author={Bayhaqy, Achmad and Sfenrianto, Sfenrianto and Nainggolan, Kaman and Kaburuan, Emil R},
  booktitle={2018 international conference on orange technologies (ICOT)},
  pages={1--6},
  year={2018},
  organization={IEEE}
}

@inproceedings{kewalramani2024optimization,
  title={Optimization of Sentiment Analysis using BERT},
  author={Kewalramani, Bhaskar and Kumar, Suresh},
  booktitle={2024 International Conference on Communication, Computer Sciences and Engineering (IC3SE)},
  pages={1000--1004},
  year={2024},
  organization={IEEE}
}

@inproceedings{joshy2022analyzing,
  title={Analyzing the performance of sentiment analysis using BERT, DistilBERT, and RoBERTa},
  author={Joshy, Archa and Sundar, Sumod},
  booktitle={2022 IEEE international power and renewable energy conference (IPRECON)},
  pages={1--6},
  year={2022},
  organization={IEEE}
}

@inproceedings{zhang2024sentiment,
  title={Sentiment analysis in the era of large language models: A reality check},
  author={Zhang, Wenxuan and Deng, Yue and Liu, Bing and Pan, Sinno and Bing, Lidong},
  booktitle={Findings of the Association for Computational Linguistics: NAACL 2024},
  pages={3881--3906},
  year={2024}
}

@inproceedings{he-etal-2015-question,
    title = "Question-Answer Driven Semantic Role Labeling: Using Natural Language to Annotate Natural Language",
    author = "He, Luheng  and
      Lewis, Mike  and
      Zettlemoyer, Luke",
    editor = "M{\`a}rquez, Llu{\'i}s  and
      Callison-Burch, Chris  and
      Su, Jian",
    booktitle = "Proceedings of the 2015 Conference on Empirical Methods in Natural Language Processing",
    month = sep,
    year = "2015",
    url = "https://aclanthology.org/D15-1076/",
    doi = "10.18653/v1/D15-1076",
    pages = "643--653"
}

@article{Youm2026RevisitingSRL,
  title     = {Revisiting Semantic Role Labeling:
Efficient Structured Inference with Dependency-Informed Analysis},
  author    = {Youm, Sangpil and Jones, Leah and Heninger, Hayden and Kempa, Edward and Dorr, Bonnie},
  journal={arXiv preprint arXiv:2605.02505},
  year      = {2026},
  arxivId = {2605.02505}
}

@inproceedings{ribeiro-etal-2016-trust,
    title = "``Why Should {I} Trust You?'': Explaining the Predictions of Any Classifier",
    author = "Ribeiro, Marco  and
      Singh, Sameer  and
      Guestrin, Carlos",
    editor = "DeNero, John  and
      Finlayson, Mark  and
      Reddy, Sravana",
    booktitle = "Proceedings of the 2016 Conference of the North {A}merican Chapter of the Association for Computational Linguistics: Demonstrations",
    month = jun,
    year = "2016",
    address = "San Diego, California",
    publisher = "Association for Computational Linguistics",
    url = "https://aclanthology.org/N16-3020/",
    doi = "10.18653/v1/N16-3020",
    pages = "97--101"
}

@inproceedings{jacovi2020towards,
  title={Towards faithfully interpretable NLP systems: How should we define and evaluate faithfulness?},
  author={Jacovi, Alon and Goldberg, Yoav},
  booktitle={Proceedings of the 58th annual meeting of the association for computational linguistics},
  pages={4198--4205},
  year={2020}
}

@inproceedings{camburu2019can,
  title={Can I Trust the Explainer? Verifying Post-hoc Explanatory Methods},
  author={Camburu, Oana-Maria and Giunchiglia, Eleonora and Foerster, Jakob and Lukasiewicz, Thomas and Blunsom, Phil},
  booktitle={NeurIPS 2019 Workshop on Safety and Robustness in Decision Making},
  year={2019}
}

\end{document}